\documentclass[lettersize,journal]{IEEEtran}
\usepackage{amsmath,amsfonts}
\usepackage{algorithmic}
\usepackage{algorithm}
\usepackage{array}
\usepackage[caption=false,font=normalsize,labelfont=sf,textfont=sf]{subfig}
\usepackage{textcomp}
\usepackage{stfloats}
\usepackage{url}
\usepackage{verbatim}
\usepackage{graphicx}
\usepackage{cite}

\usepackage{multirow}
\usepackage{booktabs}
\usepackage{color}
\usepackage{pifont}
\usepackage{makecell}

\def\Xcal{\mathcal{X}}
\def\Rbb{\mathbb{R}}
\def\Ecal{\mathcal{E}}
\def\Dcal{\mathcal{D}}
\def\Ebb{\mathbb{E}}
\def\Zcal{\mathcal{Z}}
\def\Mcal{\mathcal{M}}
\def\Fcal{\mathcal{F}}
\def\Ical{\mathcal{I}}
\def\Rcal{\mathcal{R}}
\def\Ycal{\mathcal{Y}}
\def\Pbf{\mathbf{P}}
\def\Vbf{\mathbf{V}}
\def\Ibf{\mathbf{I}}

\def\Ubf{\mathbf{U}}
\def\Nbf{\mathbf{N}}
\def\lbf{\boldsymbol{l}}

\newcommand{\best}[1]{{\textbf{#1}}}
\newcommand{\subbest}[1]{{\underline{#1}}}
\newcommand{\ann}[2]{\textcolor{red}{\\ANN\\}}

\begin{document}

\title{RGB Pre-Training Enhanced Unobservable Feature Latent Diffusion Model for Spectral Reconstruction}

\author{Keli Deng, Jie Nie, \IEEEmembership{Member,~IEEE}, and Yuntao Qian, \IEEEmembership{Senior Member,~IEEE}
        % <-this % stops a space
\thanks{This work was supported in part by the National Key Research and Development Program of China under Grant 2023YFE0204200 and in part by the National Natural Science Foundation of China under Grant 62071421. (Corresponding author: Yuntao Qian)
 }% <-this % stops a space
\thanks{Keli Deng and Yuntao Qian are with the Institute of Artificial Intelligence, College of Computer Science, Zhejiang University, Hangzhou 310027, China (e-mail: kldeng@zju.edu.cn; ytqian@zju.edu.cn).}
\thanks{
Jie Nie is with the Department of Computer Science and Technology, Ocean University of China, Qingdao 266100, China (e-mail: niejie@ouc.edu.cn).}
}

\markboth{}%
{}


\maketitle

\begin{abstract}
Spectral reconstruction (SR) is a challenging and crucial problem in computer vision and image processing that requires reconstructing hyperspectral images (HSIs) from the corresponding RGB images. A key difficulty in SR is estimating the unobservable feature, which encapsulates significant spectral information not captured by RGB imaging sensors. The solution lies in effectively constructing the spectral-spatial joint distribution conditioned on the RGB image to complement the unobservable feature. Since HSIs share a similar spatial structure with the corresponding RGB images, it is rational to capitalize on the rich spatial knowledge in RGB pre-trained models for spectral-spatial joint distribution learning. To this end, we extend the RGB pre-trained latent diffusion model (RGB-LDM) to an unobservable feature LDM (ULDM) for SR. As the RGB-LDM and its corresponding spatial autoencoder (SpaAE) already excel in spatial knowledge, the ULDM can focus on modeling spectral structure and integrating spectral and spatial knowledge. Moreover, separating the unobservable feature from the HSI reduces the redundant spectral information and empowers the ULDM to learn the joint distribution in a compact latent space. Specifically, we propose a two-stage pipeline consisting of spectral structure representation learning and spectral-spatial joint distribution learning to transform the RGB-LDM into the ULDM. In the first stage, a spectral unobservable feature autoencoder (SpeUAE) is trained to extract and compress the unobservable feature into a 3D manifold aligned with RGB space. In the second stage, the spectral and spatial structures are sequentially encoded by the SpeUAE and the SpaAE, respectively. The ULDM is then acquired to model the distribution of the coded unobservable feature with guidance from the corresponding RGB images. During inference, the ULDM estimates the unobservable feature guided by the RGB image, and then the HSI is reconstructed from the RGB image and the estimated feature. Experimental results on SR and downstream relighting tasks demonstrate that our proposed method achieves state-of-the-art performance.
\end{abstract}
\begin{IEEEkeywords}
Latent diffusion model, RGB pre-training, unobservable feature modeling, spectral reconstruction.
\end{IEEEkeywords}

\section{Introduction}
\IEEEPARstart{H}{yperspectral} images (HSIs) provide richer spectral information than RGB images, recording details across the infrared and ultraviolet spectra, which foster comprehensive analysis of materials, textures, and environmental conditions. This characteristic makes them valuable for various downstream applications, including anomaly detection, object tracking, medical imaging, and land classification \cite{zhangSurveyComputationalSpectral2022, 7920398, 9857277}. {Furthermore, the full-spectral information in HSIs facilitates more advanced image processing, such as relighting, 3D reconstruction, and depth estimation \cite{4409090, 6532304, 7045903, 9403964}. 
In the context of relighting, the goal is to restore the true color of objects by modeling the interaction between reflectance and illumination \cite{969113}. Since the spectral response function (SSF) of an RGB camera is a wide-band filter, the captured color is influenced by the light-reflectance interaction across all wavelengths within the passband. Nevertheless, the few channels in the RGB-based relighting model restrict the ability to finely model the light-reflectance interaction, leading to unsatisfactory relighting results. In contrast, the broader spectral information in HSIs allows for complete analysis of the interaction, enhancing the color fidelity of the rendered images.
}

Despite HSIs' advantages, it is demanding to capture both spectral and spatial information accurately. The high cost of acquiring HSIs and relatively low imaging quality limit their practical use. Traditional spectrometers, which gather HSIs in a spectral or spatial scanning manner, suffer from the long exposure times and the distortions caused by the sensor movement \cite{James_2007}. One way to address these issues is the snapshot compressive imaging technique, which compresses the spectral and spatial information into a single 2D measurement and then calculates the HSI through complex reconstruction processes \cite{Hu_2022_CVPR, Johnson:06, 10.1145/3130800.3130896}. Others propose to fuse a high-resolution RGB image and a low-resolution HSI into a high-resolution HSI \cite{10137388, pang2024hir, dongNoisePriorKnowledge2023}. While promising, these methods still depend on specialized and often costly equipment, limiting their accessibility and scalability.

To reduce the cost of obtaining HSIs, spectral reconstruction (SR) has emerged as a favorable solution, reconstructing HSIs from easily accessible RGB images \cite{Arad_2022_CVPR_recovery}. However, SR poses a significant challenge as an inverse problem due to the higher spectral dimensionality of HSIs compared to RGB images. This disparity introduces a great magnitude of unobservable features, which encapsulate critical spectral information filtered out by RGB imaging sensors. Without additional information, the possible unobservable feature is infinite. To obtain the optimal unobservable feature from the corresponding RGB image, it is necessary to precisely model the spectral-spatial joint distribution of unobservable features conditioned on the RGB images.

Traditional SR methods typically rely on dictionary learning with sparsity or other handcrafted constraints \cite{10.1007/978-3-319-46478-7_2, 8237766, 8481553}. Nevertheless, their effectiveness in complicated scenes comes down with the lack of representation and generalization capacity. With advancements in deep learning, convolutional neural networks (CNNs) and transformer-based methods have shown their ability to capture local and global spatial structure through 2D convolution and attention blocks \cite{9857277}. These methods employ end-to-end training on RGB-HSI pairs to learn a direct mapping from RGB images to HSIs. In addition to these discriminative methods, generative methods have emerged as powerful tools for solving RGB inverse problems \cite{li2023diffusion_survey, saharia2021image}. Among them, generative adversarial networks (GANs) and diffusion models (DMs) have recently been applied to SR \cite{10214426, liu2021hyperspectral, 10281589}, which model the spectral-spatial joint distribution conditioned on RGB images and yield realistic results from the distribution. Despite their superiority, the generative methods face two critical issues for effectively learning the joint distribution: the scarcity of RGB-HSI training pairs, which limits robustness across different scenes and illuminants, and the inherently high spectral and spatial dimensions of HSIs, which make the distribution learning particularly difficult. 

Accordingly, adapting the RGB pre-trained latent diffusion model (RGB-LDM) for unobservable feature modeling is an attractive and rational solution. First, HSIs have analogous spatial structures with the corresponding RGB images, making it possible to employ the spatial knowledge in the RGB-LDM derived from external RGB images for spectral-spatial joint distribution learning. Second, the RGB-LDM efficiently learns the distribution in a latent space spanned by the corresponding spatial autoencoder (SpaAE), motivating us to represent the spectral-spatial structure in a latent space for joint distribution learning. Third, the unobservable feature contains less information than the whole HSI, promoting low-dimensional representation and joint distribution learning.

\begin{figure*} 
    \centering
    \includegraphics[width=1\textwidth]{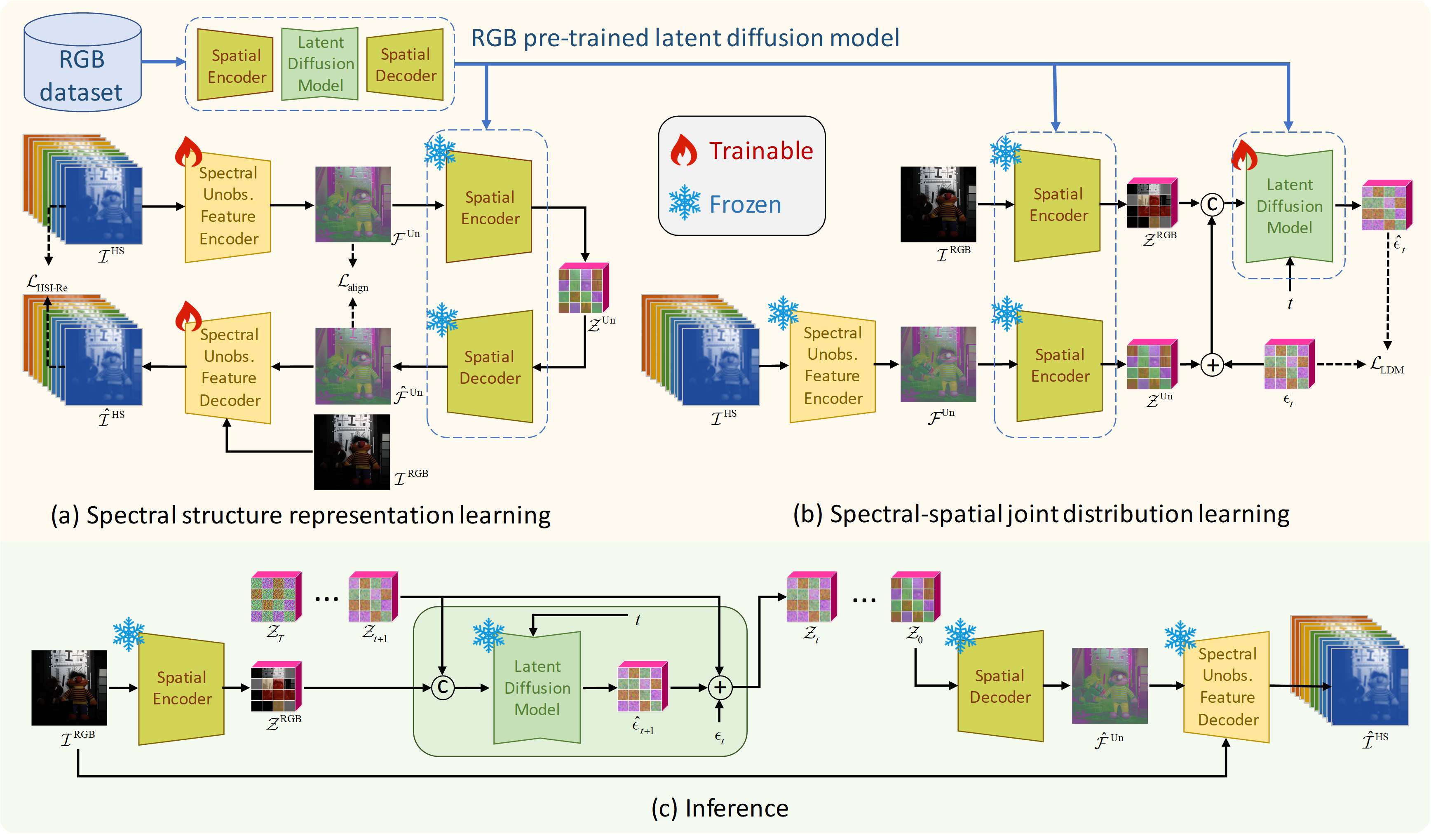} % Reduce the figure size so that it is slightly narrower than the column.
    \caption{The overall framework of the proposed method. (a) The first training stage. The unobservable feature $\Fcal^\text{Un}$ is extracted as a low-dimensional representation of the spectral structure by SpeUAE. (b) The second training stage. $\Fcal^\text{Un}$ is further encoded by the SpaAE of RGB-LDM, and then the ULDM is fine-tuned to model the coded spectral-spatial joint distribution $p(\Zcal^\text{Un}|\Zcal^\text{RGB})$. (c) The inference stage. The unobservable feature $\hat \Fcal^\text{Un}$ is estimated by the ULDM, and the HSI $\hat\Ical^\text{HS}$ is derived from the RGB image $\Ical^\text{RGB}$ and $\hat \Fcal^\text{Un}$ by SpeUAE.
    }
    \label{stage1}
\end{figure*}

In this work, we design a two-stage training pipeline to effectively convert the RGB-LDM into an unobservable feature LDM (ULDM). The first stage is spectral structure representation learning through a spectral unobservable feature autoencoder (SpeUAE). The SpeUAE comprises a spectral encoder that embeds the spectral unobservable feature from the HSI into a 3D manifold and a spectral decoder that reconstructs HSI from the RGB image and the feature. To harness the advantages in the RGB-LDM, the 3D unobservable feature manifold derived from the SpeUAE is regularized to resemble the RGB space. The second stage is spectral-spatial joint distribution learning, where the spectral and spatial structures are encoded by the SpeUAE and the SpaAE sequentially. The ULDM is initialized by the RGB-LDM and fine-tuned to exert the spatial distribution for modeling the spectral-spatial joint distribution conditioned on the corresponding RGB image. During inference, the unobservable feature is estimated by the ULDM with guidance from the RGB image, and the HSI is reconstructed by the SpeUAE from the RGB image and the estimated feature. The overall framework is illustrated in Fig. \ref{stage1}.
We evaluate the proposed method through extensive experiments on three public datasets, demonstrating its state-of-the-art performance in SR and downstream relighting based on quantitative metrics.
The main contributions of this paper are summarized as:
\begin{itemize}
    \item {The ULDM is developed to separately model the unobservable feature, which facilitates spectral structure representation learning and spectral-spatial joint distribution learning.}
    \item {Spatial similarity between RGB images and HSIs is exploited to empower the unobservable feature estimation via external and professional spatial knowledge in the RGB-LDM.
    }
    \item {A disentangled representation of spectral and spatial structures is designed to ease the difficulty of representation learning by leveraging the SpaAE of RGB-LDM.}
    \item {State-of-the-art performance is achieved on three datasets through extensive experiments, with additional verification in relighting tasks that highlight the effectiveness of the reconstructed HSIs.}
\end{itemize}
\section{Related Work}
\subsection{Diffusion Model for General Inverse Problem}
The Diffusion Model (DM) is a powerful mathematical tool that can generate high-quality images competitive to GANs \cite{NEURIPS2021_49ad23d1, NEURIPS2020_4c5bcfec, rombach2022highresolution}. It contains a forward process that gradually degrades an image to Gaussian noise so that it can draw samples from the data distribution by a learned reverse process that recovers an image from noise. Much research has found that DM is appropriate for various vision inverse problems, such as denoising, inpainting, and super-resolution \cite{li2023diffusion_survey, saharia2021image, ke2024repurposing, chung2023prompttuning}.

As the development of DM, the LDM is proposed for addressing high-resolution image generation tasks in a quantized latent space by a SpaAE \cite{rombach2022highresolution}. Within this space, the spatial structure is more efficiently managed, permitting the LDM to establish the data distribution and generate samples effectively. Based on the LDM framework, numerous high-performance RGB-LDMs have been presented, trained on internet-scale RGB images, and off-the-shelf for various vision tasks. Chung et al. \cite{chung2023prompttuning} propose a prompt tuning strategy for adapting the prompt of the RGB-LDM for vision reconstruction problems. Ke et al. \cite{ke2024repurposing} propose a fine-tuning protocol to further unlock the potential of the RGB-LDM for monocular depth estimation by keeping the latent space intact. However, these existing fine-tuning algorithms and RGB-LDMs are designed for RGB images, lacking the ability to handle the rich spectral information in HSIs.

\subsection{Spectral Reconstruction}
Traditional SR methods mainly construct handcrafted or shallow spectral features under the low-dimensional manifold assumption. Arad et al. \cite{10.1007/978-3-319-46478-7_2} explore the manifold based on dictionary learning and calculate spectral features by sparse regression. SR is addressed by mapping the sparse features to the HSI using the learned spectral dictionary. Isometric mapping algorithm and Gaussian process have also been used to represent the spectral features in the manifold \cite{8237766, 8481553}. These methods assume that the spectra can be reconstructed by merely the corresponding pixel in the RGB image and fall short of fully utilizing spatial information in the image. Additionally, their reliance on handcrafted spectral features limits the reconstruction capability in complex scenarios.

To utilize both spectral and spatial features, CNN and transformer-based methods are proposed for SR. HSCNN \cite{8265278} and HSCNN+ \cite{8575292} successfully apply the CNNs to capture the spatial feature and map it to the spectral feature by training on RGB-HSI pairs. A pixel-aware CNN is then introduced to utilize distinct reconstruction functions for different pixels, establishing the spectral feature with a flexible receptive spatial field \cite{zhang2019pixelaware}. AWAN \cite{9150664} combines the spatial attention block with the CNN block to process the long-range spatial dependency. MST++ \cite{9857277} further improves the performance by introducing spectral attention blocks to capture the spectral self-similarity.

Recently, generative methods have been studied for SR. R2HGAN \cite{liu2021hyperspectral} and HSGAN \cite{10214426} introduce the GAN to model the spectral-spatial joint distribution by adversarial training and lead to realistic HSI generation. Nevertheless, these methods can not estimate a robust joint distribution from limited RGB-HSI pairs for high-quality HSI reconstruction in complicated scenes.
R2H-CCD \cite{10281589} is a DM-based SR algorithm, which employs pixel shuffling to convert the HSI into a pseudo-RGB image, which reduces the spectral dimension and increases spatial dimension. Therefore, the SR is transformed into a spatial super-resolution problem to obtain the pseudo-RGB image by up-sampling the corresponding RGB image. However, the pixel shuffling operator introduces distortions to the spatial domain and leads to spatial dimensional exposure, creating difficulties in constructing the joint distribution. In contrast, the proposed ULDM represents the spectral-spatial structure in a robust latent space, making it powerful at distribution modeling.

\section{{Unobservable Feature in SR}}
SR problem is formulated as the inverse problem of the spectral imaging process to recover the HSI $\Ical^\text{HS}\in\Rbb^{B\times N_x\times N_y}$ from the RGB image $\Ical^{\text{HS}}\in\Rbb^{3\times N_x\times N_y}$, which is
\begin{equation}
    \label{eq:rgb2hsi}
    \Ical^\text{RGB} = \Ical^\text{HS} \times_1 \Pbf ^T
\end{equation}
where $\times_1$ is the tensor mode-1 product, $B\gg3$ is the number of wavelengths, $N_x\times N_y$ is the spatial resolution, and $\Pbf\in\Rbb^{3\times B}$ is the SSF. An intuitive idea adopted by existing generative methods is to directly learn the spectral-spatial joint distribution conditioned on the RGB image $p(\Ical^\text{HS}|\Ical^\text{RGB})$, but suffering from the high-dimensional nature of $\Ical^\text{HS}$. Thereby, we split the $\Ical^\text{HS}$ into observable and unobservable features for dimensionality reduction.

Applying the singular value decomposition (SVD) to $\Pbf$, we denote $\Vbf_0\in\Rbb^{B\times 3}$ as the right singular vectors corresponding to the zero singular values, and $\Vbf_1\in\Rbb^{B\times(B-3)}$ as the right singular vectors corresponding to the non-zero singular values. Based on the property of SVD, we can transform $x^{\text{HS}}$ into the combination of observable and unobservable features, as
\begin{align}
    \label{eq:feature space}
    \Ical^{\text{HS}} &= \Fcal^\text{Ob}\times_1 \Vbf_1^T + \Fcal^\text{Un}\times_1 \Vbf_0^T
\end{align}
where $\Fcal^\text{Ob}=\Ical^{\text{HS}}\times_1 \Vbf_1$ and $\Fcal^\text{Un}=\Ical^{\text{HS}}\times_1 \Vbf_0$ are the observable and unobservable features, respectively.
Since $\Pbf\Vbf_{0}=\mathbf{O}$, Eq. (\ref{eq:rgb2hsi}) can be rearranged to
\begin{align}
    \Ical^{\text{RGB}} 
    \label{eq:feature space with ssf}
    &= \Fcal^\text{Ob}\times_1 \Vbf_1^T\Pbf^T
\end{align}
Eq. (\ref{eq:feature space with ssf}) indicates that $\Fcal^\text{Un}$ is dropped out by the RGB camera, and any change in $\Fcal^\text{Un}$ will not affect the value of $\Ical^\text{RGB}$. Furthermore, $\Ical^{\text{RGB}}$ is equivalent to $\Fcal^\text{Ob}$, which is an exact representation of a portion of the HSI without the need for additional modification.

Due to the missing feature, the solution to the problem (\ref{eq:rgb2hsi}) is not unique, which can be formulated as 
\begin{equation}
    \label{eq:solution space}
    \hat\Ical^\text{HS} = \Ical^{\text{RGB}}\times_1 (\Vbf_1^T\Pbf^T)^{-1}\Vbf_1^T + \zeta\times_1 \Vbf_0^T
\end{equation}
where $\zeta\in\Rbb^{(B-3)\times N_x\times N_y}$ is an arbitrary tensor. To find the high-fidelity HSI, we capitalize on the spectral-spatial joint distribution to estimate the probable unobservable feature conditioned on the RGB image. Formally, we can let $\zeta = \Fcal^\text{Un} \sim p(\Fcal^\text{Un}|\Ical^\text{RGB})$ to derive a well estimated HSI from Eq. (\ref{eq:solution space}).

Nevertheless, the dimension of original $\Fcal^\text{Un}$ is still high, raising the problem for distribution learning. Accordingly, we utilize the SpeUAE to compress $\Fcal^\text{Un}$ and reconstruct the HSI taking the place of Eq. (\ref{eq:solution space}). Moreover, the SpaAE of RGB-LDM is employed to code the spatial structure in $\Fcal^\text{Un}$ and finally obtain $\Zcal^\text{Un}$, which has less dimensionality and information than $\Ical^\text{HS}$. $\Ical^\text{RGB}$ is also coded to $\Zcal^\text{RGB}$ by the SpaAE. Based on these representations, the problem of learning $p(\Ical^\text{HS}|\Ical^\text{RGB})$ is addressed by learning the embedded joint distribution conditioned on RGB images $p(\Zcal^\text{Un}|\Zcal^\text{RGB})$ with ULDM.

\section{Extending RGB-LDM to ULDM}
In this section, we introduce the two-stage training pipeline to extend the RGB-LDM to the ULDM, which models the spectral-spatial joint distribution of the unobservable feature. Furthermore, an analysis of the estimated unobservable feature by the ULDM is provided.
\subsection{Spectral Structure Representation Learning}
In this stage, the goal is to represent the spectral structure to facilitate the subsequent joint distribution learning. To this end, we train the SpeUAE with an HSI reconstruction loss to extract the unobservable features from HSIs and encode them into a 3D manifold. It is followed by the SpaAE with an alignment loss as regularization, ensuring the proper embedding of spatial structure in the next stage.

The low-rank property of the spectrum allows for compressing the spectral structure into a low-dimensional manifold \cite{Nie_2018_CVPR}. However, the traditional spectral autoencoder attempts to represent the entire spectrum at once, overlooking the fact that the RGB image itself effectively represents part of the HSI. In contrast, SpeUAE separately encodes the unobservable feature from the HSI, yielding a high accuracy in spectral structure representation under the same dimensionality.
Specifically, the compressed unobservable feature $\Fcal^\text{Un}\in\Rbb^{3\times N_x\times N_y}$ is obtained by
\begin{eqnarray}
    \label{eq:spectral encoder}
    \Fcal^\text{Un} = \Ecal_\theta(\Ical^{\text{HS}})
\end{eqnarray}
where $\Ecal_\theta$ is the encoder with parameters $\theta$.
According to Eq. (\ref{eq:solution space}), the HSI is uniquely reconstructed by the RGB image and the unobservable feature, which can be formulated as
\begin{eqnarray}
    \label{eq:spectral decoder}
    \hat\Xcal ^{\text{HS}} = \Dcal_\vartheta(\Fcal^\text{Un}, \Ical^{\text{RGB}})
\end{eqnarray}
where $\Dcal_\vartheta$ is the decoder with parameters $\vartheta$. Since $\Ical^\text{RGB}$ explicitly participates in the decoding process, $\Ecal_\theta$ does not need to redundantly handle the RGB image and can be dedicated to the unobservable feature. This will be further validated in the ablation study.

\begin{figure}
    \centering
    \includegraphics[width=0.98\columnwidth]{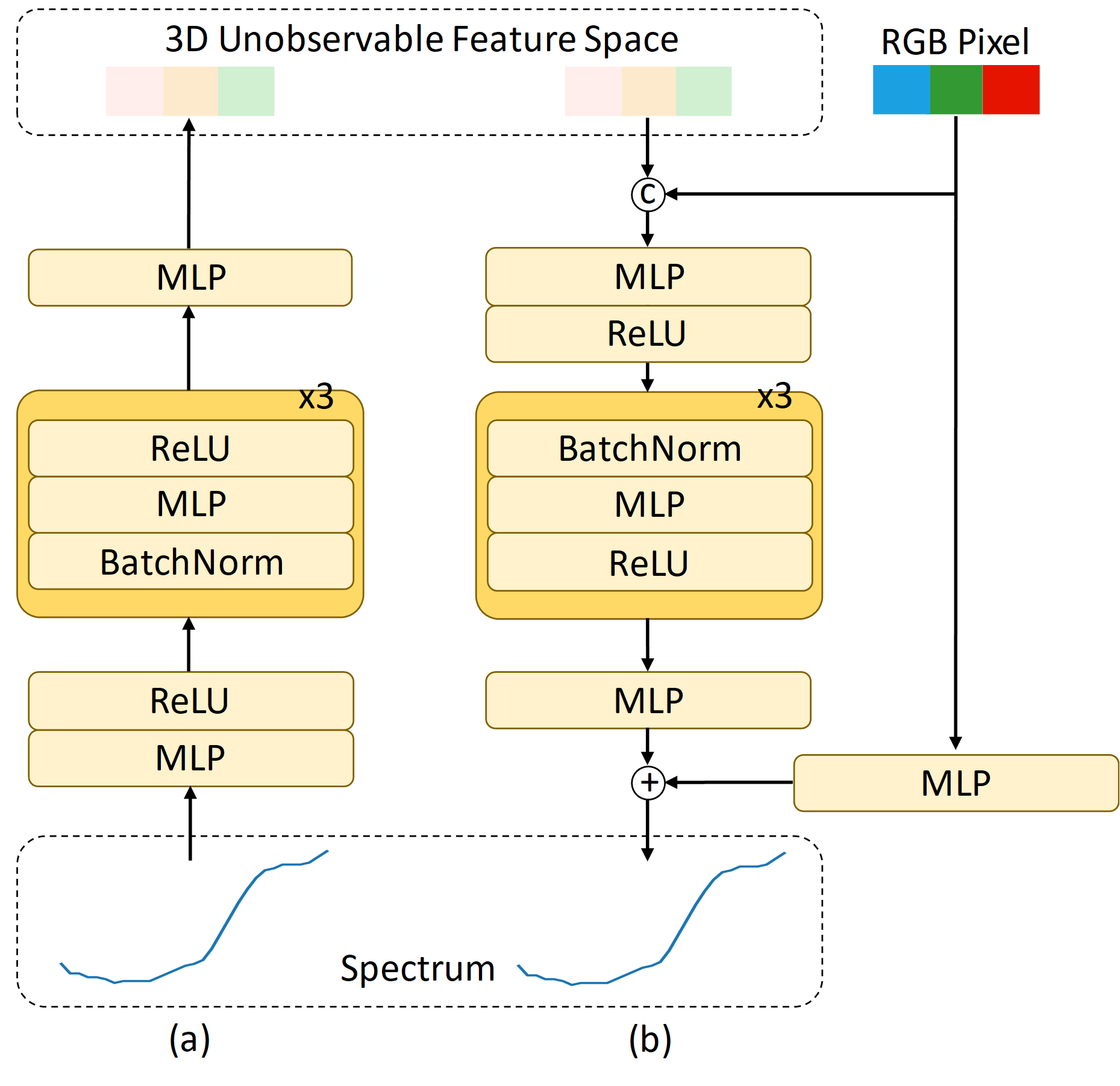} % Reduce the figure size so that it is slightly narrower than the column.
    \caption{The architecture of the SpeUAE. (a) The spectral unobservable feature encoder $\Ecal_\theta$. (b) The spectral unobservable feature decoder $\Dcal_\vartheta$.}
    \label{spatial autoencoder}
\end{figure}

The architecture of $\Ecal_\theta$ and $\Dcal_\vartheta$ are shown in Fig. \ref{spatial autoencoder}, primarily implemented by stacking several MLP layers to compress and decompress the unobservable feature. A skipping connection is added to the RGB image in $\Dcal_\vartheta$ as the RGB image can linearly represent a portion of the HSI. According to the definition of the unobservable feature, $\Ecal_\theta$ and $\Dcal_\vartheta$ only manipulate the spectral structure pixel-by-pixel.

The feature space learned by an autoencoder is not unique \cite{Wei20237589}. This flexibility allows us to regularize the 3D unobservable feature manifold to resemble the RGB space, ensuring that the spatial structure in $\Fcal^\text{Un}$ can be directly managed by the RGB-LDM. However, it is challenging to measure the similarity between the 3D manifold and the RGB space. Instead, we introduce the alignment loss $\mathcal{L}_{\text{align}}$ to minimize the errors caused by the SpaAE of RGB-LDM.
Specifically, the spatial encoder $\Ecal_\text{spatial}$ transforms $\Fcal^\text{Un}$ into a latent code $\Zcal^\text{Un}$ as
\begin{eqnarray}
    \label{eq:vq encoder}
    \Zcal^\text{Un} = \Ecal_\text{spatial}(\Fcal^\text{Un})
\end{eqnarray}
where the latent code $\Zcal^\text{Un}$ compresses the spatial structure of $\Fcal^\text{Un}$. Subsequently, the spatial decoder $\Dcal_\text{spatial}$ reconstructs the unobservable feature $\hat\Fcal^\text{Un}$ as
\begin{eqnarray}
    \label{eq:vq decoder}
    \hat\Fcal^\text{Un} = \Dcal_\text{spatial}(\Zcal^\text{Un})
\end{eqnarray}
$\Ecal_\text{spatial}$ and $\Dcal_\text{spatial}$ leverage the spatial low-rank property of images, learning from internet-scale RGB images to represent the spatial structure into a compact latent space.
$\mathcal{L}_{\text{align}}$ is used to minimize the errors in $\hat\Fcal^\text{Un}$ caused by the SpaAE, which is
\begin{eqnarray}
    \label{eq:alignment loss}
    \mathcal{L}_{\text{align}} = \Ebb_{\Fcal^\text{Un}}\|\hat\Fcal^\text{Un}-\Fcal^\text{Un}\|_2^2
\end{eqnarray}
During this training stage, we fix the SpaAE's parameters to preserve its original spatial knowledge derived from RGB images. By optimizing $\mathcal{L}_\text{align}$, we can guarantee that $\Dcal_\text{spatial}(\Ecal_\text{spatial}(\Fcal^\text{Un}))\simeq\Fcal^\text{Un}$, indicating that $\Fcal^\text{Un}$ shares the same latent space with RGB images from the perspective of $\Ecal_\text{spatial}$ and $\Dcal_\text{spatial}$. Thereby, the 3D unobservable feature manifold is well aligned with the RGB space in the context of the SpaAE.
% Additionally, the intact latent space enables us to extend the spatial distribution in the RGB-LDM to spectral-spatial joint distribution in the next stage.

The HSI reconstruction loss $\mathcal{L}_{\text{HSI-Re}}$ is introduced to confirm that the compressed unobservable feature encapsulates essential spectral information to reconstruct the HSI together with the RGB image, which is
\begin{align}
    \label{eq:hsi reconstruction loss}
    \mathcal{L}_{\text{HSI-Re}} = \Ebb_{(\Ical^\text{HS},\Ical^\text{RGB})}&\|\Ical^\text{HS} - \hat\Ical^\text{HS}\|_2^2
\end{align}

Eventually, the total loss function in this stage is obtained by the weighted sum of $\mathcal{L}_{\text{HSI-Re}}$ and $\mathcal{L}_{\text{align}}$, which is
\begin{eqnarray}
    \label{eq:total loss}
    \mathcal{L}_\text{SpeUAE} = \mathcal{L}_{\text{HSI-Re}} + \lambda\mathcal{L}_\text{align}
\end{eqnarray}
where $\lambda$ is the weight parameter.

As the spectral structure is encoded into a manifold aligned with the RGB space by the SpeUAE, the spatial structure can be encoded by the SpaAE precisely. This decoupled design efficiently creates a robust low-dimensional latent space to represent the spectral-spatial structure, which is beneficial for the subsequent joint distribution learning. 

\subsection{Spectral-Spatial Joint Distribution Learning}
In this stage, we aim to construct the spectral-spatial joint distribution conditioned on the RGB image for unobservable feature modeling. Specifically, the ULDM is derived from fine-tuning the RGB-LDM to recover unobservable features in the latent space from noisy features guided by RGB images.

Initially, the unobservable feature is calculated by the SpeUAE and then embedded by the SpaAE as
\begin{eqnarray}
    \label{eq:vq encoder hsi}
    \Zcal^\text{Un} = \Ecal_\text{spatial}(\Fcal^\text{Un}) = \Ecal_\text{spatial}(\Ecal_\theta(\Ical^\text{HS}))
\end{eqnarray}
The RGB image is also embedded by the SpaAE as
\begin{eqnarray}
    \label{eq:vq decoder rgb}
    \Zcal^\text{RGB} = \Ecal_\text{spatial}(\Ical^\text{RGB})
\end{eqnarray}
where $\Zcal^\text{Un}$ and $\Zcal^\text{RGB}$ are the latent codes of the unobservable feature and the RGB image, respectively. From Eq. (\ref{eq:vq encoder hsi}) and Eq. (\ref{eq:vq decoder rgb}), we can see that $\Fcal^\text{Un}$ and $\Ical^\text{RGB}$ share the same latent supported by the $\Ecal_\text{spatial}$. This alignment makes it efficient to establish the relationship between the unobservable feature and the RGB image in the latent space by the ULDM.

Specifically, in the forward process, the given target sample $\Zcal_0:=\Zcal^\text{Un}$ is gradually corrupted with Gaussian noise. At step $t$, the noised sample can be obtained by
\begin{eqnarray}
    \label{eq:corrupted sample}
    \Zcal_t = \sqrt{\bar\alpha_t}\Zcal_{0} + \sqrt{1-\bar\alpha_t}\epsilon
\end{eqnarray}
where each element of $\epsilon$ is sampled from a standard Gaussian distribution, $\bar\alpha_t:=\prod_{s=0}^t1-\beta_s$, $\{\beta_s\}_{s=0}^T$ are hyperparameters controlling the noise level, and $T$ is the number of diffusion steps. For learning to reverse the process to gradually remove the noise from the noisy sample $\Zcal_t$ and obtain $\Zcal_{t-1}$, a conditional denoising model $\epsilon_\phi$ is employed, with $\phi$ being the model's parameters. During this process, the joint distribution is learned by denoising $\Zcal_t$ conditioned on the spatial information in $\Zcal^\text{RGB}$. To improve convergence and generalization, the RGB-LDM is used as the initialization of $\epsilon_\phi$, which provides a robust spatial distribution for joint distribution learning.

Using the re-parameterization trick \cite{NEURIPS2020_4c5bcfec}, $\epsilon_\phi$ can be fine-tuned by minimizing the distance between the injected noise $\epsilon$ and the estimated noise $\hat\epsilon:=\epsilon_\phi(\Zcal_t;\Zcal^\text{RGB}, t)$, which is formulated as
\begin{eqnarray}
    \label{eq:loss function}
    \mathcal{L}_\text{LDM} = \Ebb_{(\Zcal_0,\Zcal^\text{RGB}),\epsilon, t}\|\epsilon-\hat\epsilon\|_2^2
\end{eqnarray}
Since $\epsilon_\phi$ is already pre-trained in the latent space of $\Zcal^\text{RGB}$ spanned by $\Ecal_\text{spatial}$, it can be efficiently transformed to learn the distribution of $\Zcal^\text{Un}$ which is well-aligned with $\Zcal^\text{RGB}$ in the same latent space.

After training the SpeUAE and ULDM, we can estimate the unobservable feature given an RGB image. First, the RGB image $\Ical^\text{RGB}$ is embedded into the latent space to obtain $\Zcal^\text{RGB}$ by spatial encoder $\Ecal_\text{spatial}$. Within this latent space, the $\Zcal^\text{Un}\sim p(\Zcal^\text{Un}|\Zcal^\text{RGB})$ is then estimated by the LDM through a series of denoising steps guided by the RGB image. Subsequently, the unobservable feature $\Fcal^\text{Un}$ is derived from $\Zcal^\text{Un}$ by spatial decoder $\Dcal_\text{spatial}$. $\hat\Ical^\text{HS}$ is finally reconstructed from the $\Ical^\text{RGB}$ and $\Fcal^\text{Un}$ by the spectral unobservable feature decoder $\Dcal_\vartheta$.

\subsection{Analysis of the ULDM}
\label{sec:Analysis of the Estimated Unobservable Feature}
In this section, we analyze the estimated unobservable feature and the reconstructed HSI from the ULDM under the low-dimensional linear manifold assumption to verify the effectiveness of the proposed method.

We start by analyzing the property of the theoretical optimal denoising model $\epsilon_{\phi^\star}$ derived from Eq. (\ref{eq:loss function}). According to Tweedie's formulation \cite{dou2024diffusion} and the forward process in Eq. (\ref{eq:corrupted sample}), we have
\begin{align}
    \label{eq:optimal denoising}
    \hat\epsilon_{\phi^\star} &= \frac{1}{\sqrt{1-\bar\alpha_t}}(\Zcal_t-\sqrt{\bar\alpha_t} \Zcal_{0|t})
\end{align}
where $\Zcal_{0|t}:=\mathbb{E}_{\Zcal_0\sim p(\Zcal_0|\Zcal_t, \Zcal^\text{RGB})}[\Zcal_0]$ is the denoising result obtained at step $t$. Applying the Bayes' rule to $\Zcal_{0|t}$, we have
\begin{align}
    \Zcal_{0|t} 
    % &\propto \int \Zcal_0 p(\Zcal_t|\Zcal_0) p(\Zcal_0|\Zcal^\text{RGB})d\Zcal_0 \\
    &\propto \int \Zcal_0 p(\Zcal_t|\Zcal_0) p(\Zcal_0) p(\Zcal^\text{RGB}|\Zcal_0)d\Zcal_0
    \label{eq:z0}
\end{align}
where $p(\Zcal_0) := p(\Zcal^\text{Un})$ is the prior distribution of the unobservable feature, and $ p(\Zcal^\text{RGB}|\Zcal_0)$ can be calculated by the spectral imaging process in Eq. (\ref{eq:rgb2hsi}). Therefore, Eq. (\ref{eq:z0}) can be further simplified as
\begin{align}
    \label{eq:z0_1}
   & \Zcal_{0|t} \propto \int_{\Mcal_0} \Zcal_0 p(\Zcal_t|\Zcal_0) p(\Zcal_0) d\Zcal_0 \\
   &s.t.\ \ \Mcal_0 = \{\Zcal_0|\Dcal(\Zcal_0, \Zcal^\text{RGB})\times_1 \Pbf = \Dcal(\Zcal^\text{RGB})\} 
\end{align}
where $\Mcal_0$ is the manifold supported by the physical constraint. For simplicity, we note $\Dcal$ as the decoder for both $\Zcal_0$ and $\Zcal^\text{RGB}$.

Based on the low-rank property of the spectrum \cite{10.1007/978-3-319-46478-7_2}, we can assume that the spectra in the HSI follow a low-dimensional linear manifold. Since the manifold may vary across different images \cite{8481553}, we assume the prior distribution of $\Zcal_0$ consists of $M$ linear manifolds $\Mcal_i$,
\begin{align}
    \label{eq:manifold assumption}
    \Mcal_i = \{\Ecal (\Xcal \times_1 \Ubf_i^T) | \Xcal \in \mathbb R^{L_i\times N_x\times N_y}\}
\end{align}
where $L_i$ is the dimension of the manifold, $\Xcal$ is the feature in the manifold, $\Ubf_i$ is the corresponding spectral base, and $i = 1,2,...,M$. We note $\Ecal(\cdot)$ as the encoder for simplicity.
% Different scenarios may have different $\Mcal_i$.
According to the sparsity of the HSI \cite{8340224}, it is reasonable to assume that only a particular spectral manifold corresponds to a certain scene, indicating that $\Mcal_0$ merely intersects with $\Mcal_k$. Therefore, Eq. (\ref{eq:z0_1}) can be further simplified as
\begin{align}
    \label{eq:z0_2}
   \Zcal_{0|t} \propto \int_{\Mcal_0\cap\Mcal_k} \Zcal_0 p(\Zcal_t|\Zcal_0) p(\Zcal_0) d\Zcal_0 
\end{align}
When $t\rightarrow 0$, we have $p(\Zcal_t|\Zcal_0) \rightarrow \delta (\Zcal_t - \Zcal_0)$. Therefore, the final estimated unobservable feature by LDM can be derived from Eq. (\ref{eq:z0_2}) as
\begin{align}
    \lim_{t\rightarrow 0}\Zcal_{0|t} &= \int_{\Mcal_0\cap\Mcal_k} \Zcal_0 p(\Zcal_0) \delta (\Zcal_t - \Zcal_0) d\Zcal_0
    &= \Zcal_0
    \label{eq:final}
\end{align}
where the obtained $\Zcal_0$ will follow the distribution $p(\Zcal^\text{Un})$ and must satisfy $\Zcal_0 \in \Mcal_0 \cap\Mcal_k$. Otherwise, the integral in Eq. (\ref{eq:final}) will equal 0. 

Solving $\Mcal_0 \cap\Mcal_k$ we can obtain the reconstructed HSI by the ULDM.
For the case $L_k = 3$, the reconstructed HSI from $\Zcal_0$ can be formulated as
\begin{align}
    \hat\Ical^\text{HS} &= \Ical^\text{RGB} \times (\Ubf_k^T \Pbf^T)^{-1} \Ubf_k^T = \Ical^\text{HS}
\end{align}
The HSI is perfectly reconstructed.
For the case $L_k > 3$, $\hat\Ical^\text{HS}$ is not unique, the solution space can be formulated as
\begin{align}
    % \label{eq:solution space}
    \mathcal{S} :=\left\{ \Ical^\text{RGB}\times_1 (\Ubf_k^T \Pbf^T)^\dag \Ubf_k^T + \zeta\times_1 \Nbf^T\Ubf_k^T|\zeta   \right\}
\end{align}
where $\zeta\in\Rbb ^{(L_k-3)\times N_x\times N_y}$ is an arbitrary tensor, and $\Nbf$ is the null space of $\Pbf\Ubf_k$.
ULDM could find a high-fidelity solution in $\mathcal{S}$ with probable unobservable feature $\Zcal^\text{Un} \sim p(\Zcal^\text{Un}|\Zcal^\text{RGB})$. This result indicates the feasibility and the effectiveness of the proposed method.

\section{Experiments}
In this section, we detail the implementation of our proposed method and elaborate the evaluation results on three public datasets. Additionally, we analyze the influence of some important factors.
\begin{table*}[t]
  \caption{Quantitative comparison of the proposed method with state-of-the-art SR methods on the CAVE, ICVL, and NTIRE22 datasets. The best results are \best{bold}, and the second-best results are \subbest{underscored}.}
    \centering
    \begin{tabular}{c|c|ccc|ccc|ccc}
    \toprule
    \multirow{2}{*}{Method}    &\multirow{2}{*}{Architecture}& \multicolumn{3}{c|}{CAVE} & \multicolumn{3}{c|}{ICVL} & \multicolumn{3}{c}{NTIRE22} \\
    && PSNR & SSIM & SAM & PSNR & SSIM & SAM & PSNR & SSIM & SAM \\
    \midrule
    HSCNN+       &CNN& 32.77 & 0.9727 & 11.30 & 42.79 & 0.9933 & 2.155 & 32.02 & 0.9431 & 5.102 \\
    HDNet   &CNN+Attention& 34.05 & 0.9742 & 11.11 & 46.44 & 0.9981 & 1.345 & 40.02 & 0.9832 & 4.123 \\
    MPRNet     &CNN+Attention& 34.00 & 0.9782 & 10.20 & 45.97 & 0.9980 & 1.386 & 39.36 & 0.9830 & 4.392 \\
    MST         &Transformer& \subbest{35.19} & 0.9780 & 10.65 & 47.34 & 0.9986 & 1.250 & 40.39 & 0.9863 & \subbest{3.922} \\
    MST++        &Transformer& \best{35.20} & \subbest{0.9841} & \subbest{9.05} & \subbest{47.50} & \subbest{0.9987} & \subbest{1.242} & 41.16 & 0.9870 & \best{3.691} \\
    \midrule
    R2HGAN  &GAN& 33.51 & 0.9773 & 10.87 & 45.27 & 0.9979 & 1.453 & \subbest{41.18} & \subbest{0.9882} & 4.310 \\
    R2H-CCD     &DM& 33.24 & 0.9284 & 23.52 & 30.17 & 0.9599 & 10.88 & 27.97 & 0.5926 & 24.01 \\
    Ours                       &LDM& 35.05 & \best{0.9844} & \best{8.74} & \best{47.94} & \best{0.9992} & \best{1.176} & \best{42.26} & \best{0.9888} & 4.355 \\
    \bottomrule
    \end{tabular}
    \label{tab:results}
\end{table*}
\begin{figure*}
    \centering
    \includegraphics[width=1\textwidth]{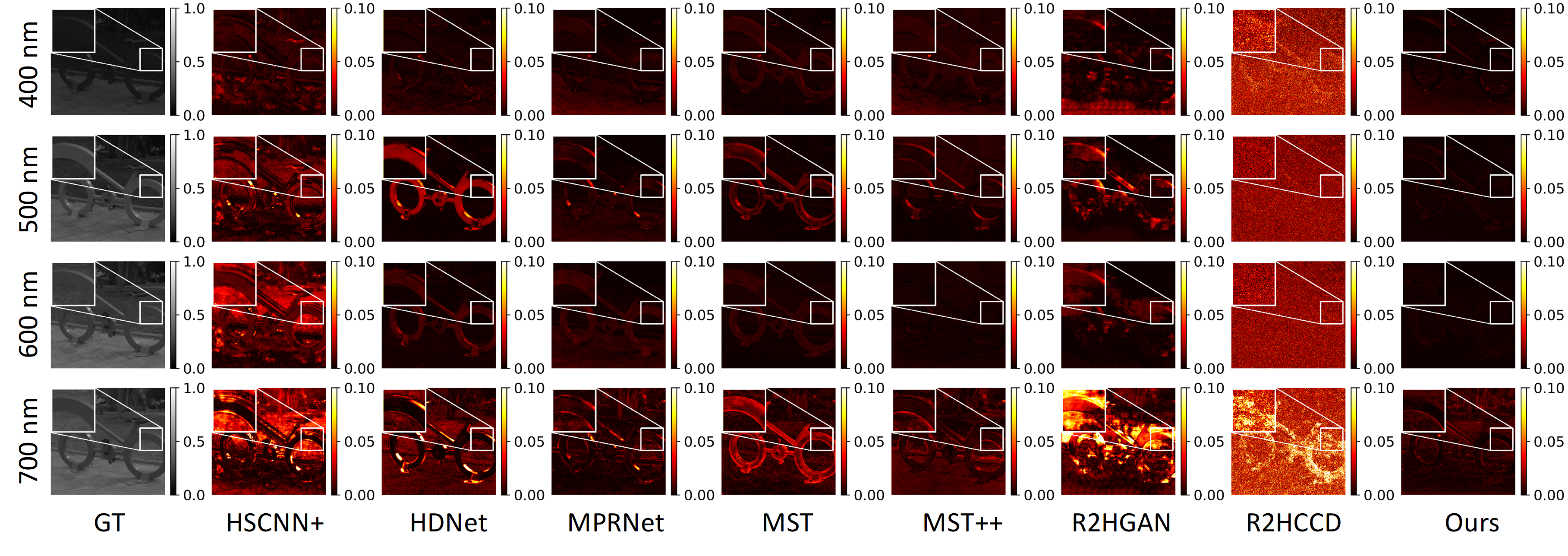} % Reduce the figure size so that it is slightly narrower than the column.
    \caption{Illustration of error maps of SR results on the NTIRE22 dataset. Four different wavelengths are shown, including 400 nm, 500 nm, 600 nm, and 700 nm. The first column is the ground truth, and the rest are the error maps of the proposed method and alternative methods.
    }
    \label{fig:visual}
\end{figure*}
\begin{figure}
    \centering
    \includegraphics[width=1\columnwidth]{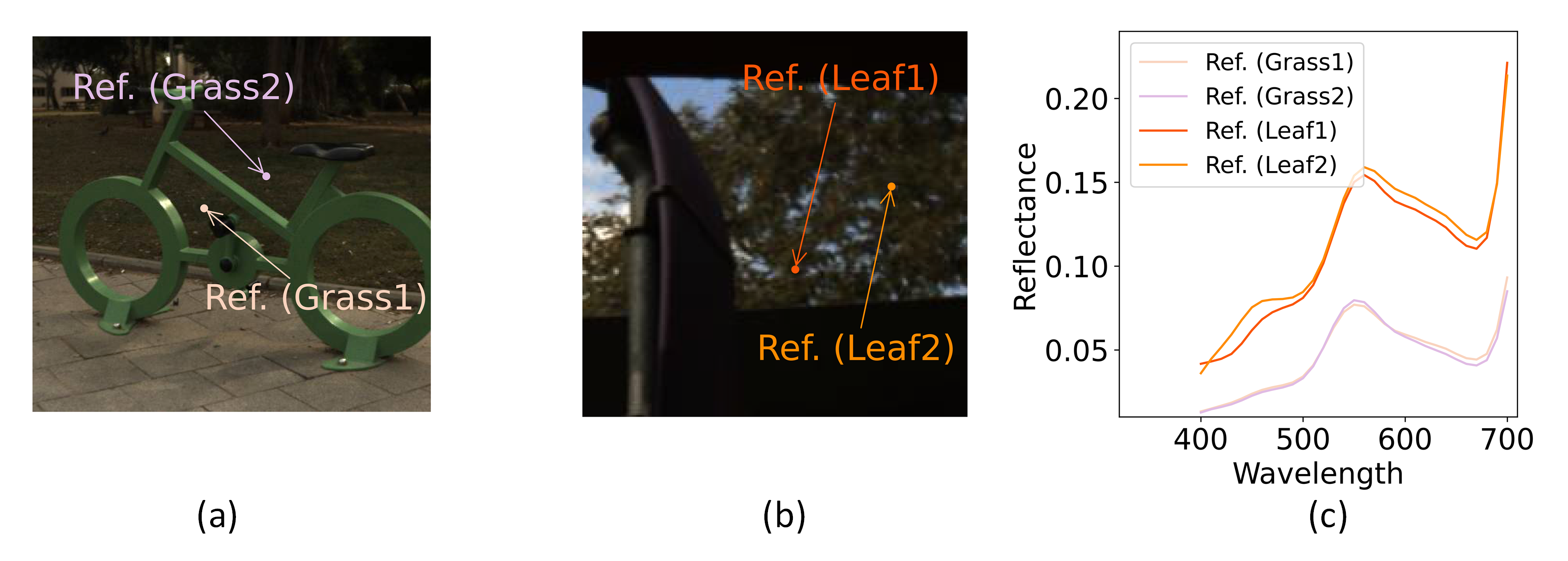} % Reduce the figure size so that it is slightly narrower than the column.
    \caption{Illustration of RGB images and the reference spectra. (a) RGB image of the grass. (b) RGB image of the leaf. (c) The reference spectra of grass and leaf.}
    \label{fig:spectral}
\end{figure}

\subsection{Implementation}
Our method is implemented using PyTorch and leverages Stable Diffusion 2.1-v \cite{rombach2022highresolution} as the RGB-LDM. During spectral structure representation, the spectral encoder and decoder of SpeUAE have a hidden dimension of 64. We start the training with an initial learning rate of $4\times 10^{-4}$. The weight parameter $\lambda$ is set to 0.1. Optimization of the encoder and decoder's parameters is carried out using the Adam optimizer over 100k iterations. The training images are cropped into $128\times128$ with a stride of 64, and a batch size of 20 is utilized. For the joint distribution learning of ULDM, the number of diffusion steps $T$ is set to 1000, and the hyperparameters $\{\beta_s\}_{s=1}^T$ are set as the same as DDPM \cite{NEURIPS2020_4c5bcfec}. The initial learning rate is set to $1.5\times 10^{-5}$. The training of ULDM is performed by the Adam optimizer over 300 epochs on 6 V100 GPUs with a batch size of 1. Data augmentation is applied, involving random rotation and flipping. During inference, DDIM sampler \cite{song2020denoising} is used with 20 re-spaced steps for accelerated sampling.
\subsection{Evaluations}
\subsubsection{Datasets} 
Three public HSI datasets are selected for evaluation, including the CAVE dataset \cite{CAVE}, ICVL dataset \cite{10.1007/978-3-319-46478-7_2}, and NTIRE 2022 Spectral Reconstruction Challenge (NTIRE22) dataset \cite{Arad_2022_CVPR_recovery}. The CAVE dataset comprises 32 HSIs with $512\times512$ spatial resolution and 31 wavelengths, where 26 HSIs are used for training and the remainder for testing. The ICVL dataset includes 201 scenes, each scene containing a 31-wavelength HSI with a spatial resolution of $1392\times1300$. We randomly select 107 HSIs for training and 30 HSIs for testing. The NTIRE22 dataset contains 1000 HSIs with $482\times512$ spatial resolution and 31 wavelengths, split into 900 for training, 50 for validation, and 50 for testing. On NTIRE22, we report the results on the validation set since the testing HSIs are not publicly available. 
The wavelengths of all datasets are range from 400 nm to 700 nm. For the CAVE and ICVL datasets, the corresponding RGB images are generated by the SSF of a Nikon D700 camera. For the NTIRE22, RGB images are generated by the SSF provided by the challenge organizers.
\subsubsection{Alternative Methods} 
{Seven} state-of-the-art methods are chosen as alternative methods for comparison,
covering HSCNN+ \cite{8575292}, HDNet \cite{Hu_2022_CVPR}, MPRNet \cite{9577298}, MST \cite{9878871}, MST++ \cite{9857277}, R2H-CCD \cite{10281589}, and R2HGAN \cite{liu2021hyperspectral}.
Among these methods, R2HGAN and R2H-CCD are generative SR methods, MST++ is transformer-based SR method, HSCNN+ is CNN-based SR methods, MPRNet is a general image reconstruction method, and MST and HDNet are two HSI reconstruction methods. Specifically, HSCNN+ and MST++ are the winners of the NTIRE 2018 and 2022 Spectral Reconstruction Challenges, respectively. For MPRNet, the number of output channels is set to the number of wavelengths. For MST and HDNet, the number of input channels is set to 3. Other hyperparameters of these alternative methods are set referring to the original papers. {To achieve a fair and statistically reliable comparison, for stochastic SR methods, we sample 10 times and average these samples to obtain the final estimated HSI.}
\subsubsection{Evaluation Metrics} 
Three widely used metrics are employed for quantitative evaluation, incorporating peak signal-to-noise ratio (PSNR), structural similarity index (SSIM), and spectral angle mapper (SAM). PSNR measures the quality of the reconstructed HSI. SSIM evaluates the spatial similarity between the reconstructed HSI and the ground truth. SAM measures the spectral similarity between the reconstructed HSI and the ground truth. PSNR and SSIM are general metrics for image quality assessment, while SAM is specific to HSIs and provides a more targeted evaluation of spectral accuracy. SAM is computed by averaging the spectral angle (SA) between the estimated and reference spectra, which is
\begin{align}
    \text{SAM}(\Ibf, \hat \Ibf) &= \frac{1}{N}\sum_{i=1}^N\text{SA}(\Ibf_i, \hat \Ibf_i) \\
    &= \frac{1}{N}\sum_{i=1}^N\arccos\left(\frac{\Ibf_i\cdot\hat \Ibf_i}{\|\Ibf_i\|\|\hat \Ibf_i\|}\right)
\end{align}
where $\Ibf$ and $\hat \Ibf$ are the flattened estimated and reference HSIs, respectively, and $N$ is the number of pixels. 
Higher PSNR and SSIM values, along with lower SA and SAM values, indicate superior reconstruction performance.
\subsubsection{Results}
The average reconstruction performances on the CAVE, ICVL, and NTIRE22 datasets are elucidated in Table \ref{tab:results}. Our method consistently achieves robust performance compared to state-of-the-art methods. Notably, our method illustrates the best reconstruction performance on the ICVL dataset across all metrics. On the CAVE and NTIRE22 datasets, our method also exhibits competitive performance across most metrics compared with alternative methods. Moreover, among generative SR methods, our approach outperforms others in almost all the metrics and datasets. R2H-CCD has poor performances across all the datasets, as its pixel shuffling operator distorts the spatial domain and leads to the dimension curse. 

The error maps, which visually represent the reconstruction errors across different wavelengths, are shown in Fig. \ref{fig:visual} for the NTIRE22 dataset. More visual results for the CAVE and ICVL datasets are provided in the Supplementary Material. It is obvious that HSCNN+ and HDNet perform poorly in reconstructing the spatial structures in most wavelengths. MPRNet and MST yield better results but still exhibit limitations in recovering high-frequency information, such as the edges of the bike. For R2HGAN, noticeable artifacts are generated in the spatial domain, probably due to the limited capability of spatial representation. MST++ achieves satisfactory performance but is still inferior to our method in reconstructing fine details. Additionally, R2H-CCD suffers from significant noise in the spatial domain, likely caused by distortion from pixel shuffling. Overall, our method consistently outperforms the others, particularly in preserving fine details and minimizing artifacts, demonstrating the effectiveness of our approach.

\begin{figure*}
    \centering
    \includegraphics[width=0.98\textwidth]{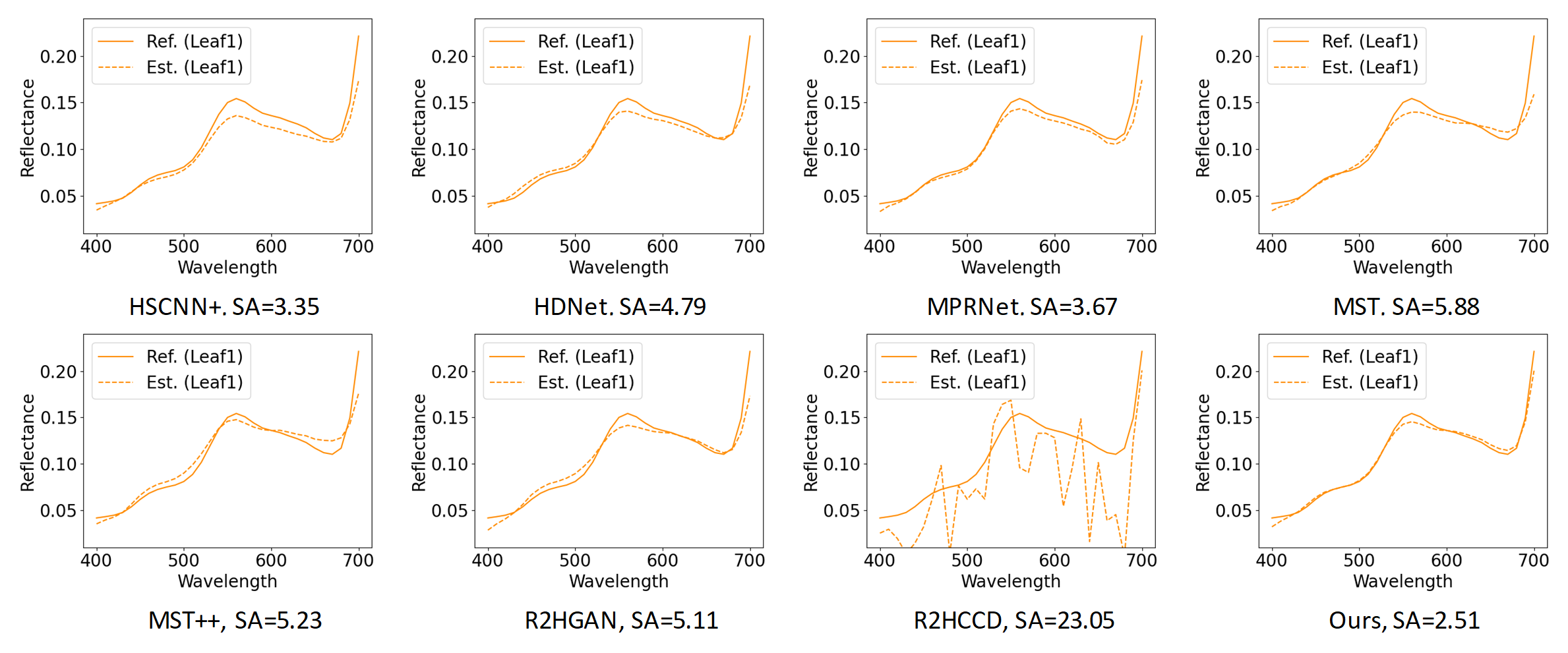} % Reduce the figure size so that it is slightly narrower than the column.
    \caption{The estimated and reference spectra of leaf. The corresponding SA metrics are calculated and presented in the figure.}
    \label{fig:estimate spectra leaf}
\end{figure*}

The capability of distinguishing between different materials with similar RGB colors is also evaluated. Two spectra of grass and leaf are selected on the NTIRE22 dataset, with their corresponding RGB images shown in Fig. \ref{fig:spectral} (a) and (b), respectively. The reference spectra are shown in Fig. \ref{fig:spectral} (c). Materials of the same type exhibit significant similarity in their spectral characteristics. Otherwise, they are distinct, even though these materials may have similar RGB colors. We quantify and visualize the SR performance using the SA metric, with results shown in Fig. \ref{fig:estimate spectra leaf} and Fig. \ref{fig:estimate spectra leaf}. Our method effectively captures these natures, recognizing the subtle differences in RGB images and estimating the corresponding spectra accurately. 

\begin{figure*}
    \centering
    \includegraphics[width=0.98\textwidth]{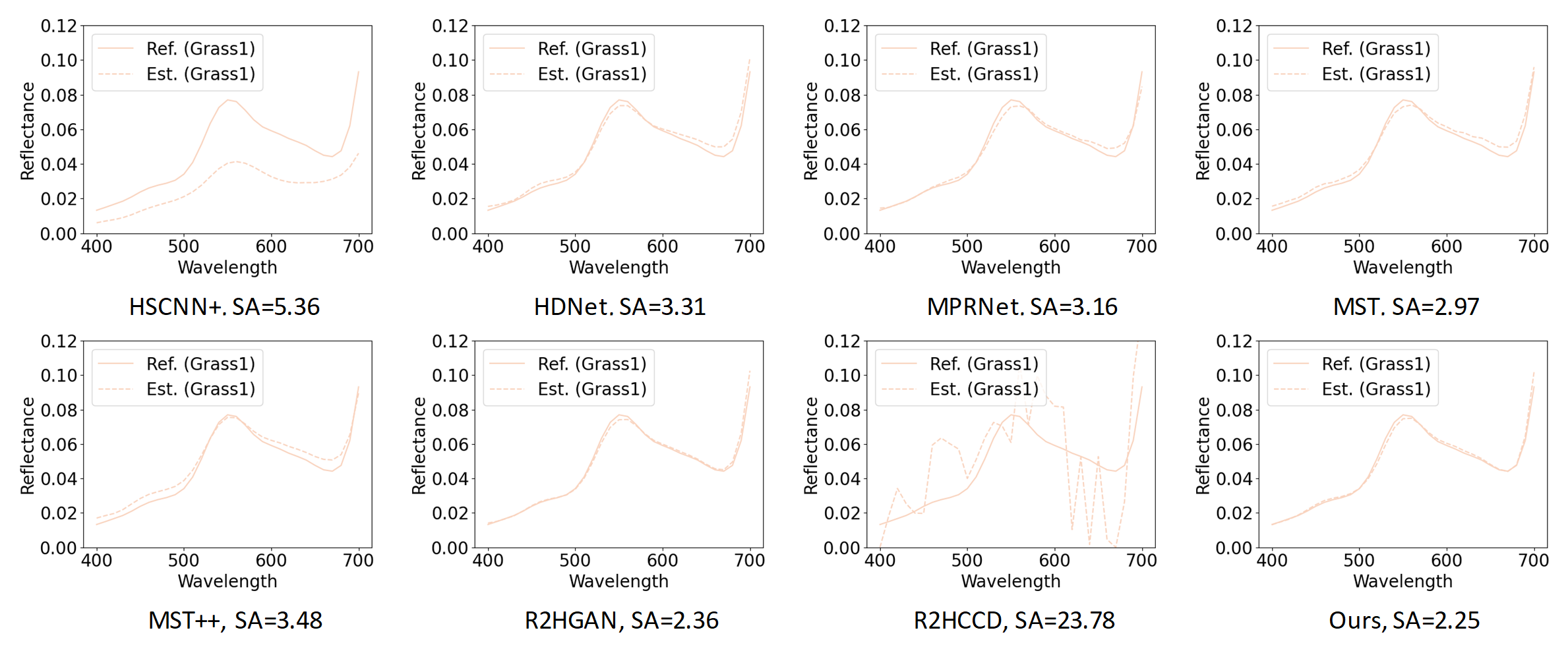} % Reduce the figure size so that it is slightly narrower than the column.
    \caption{The estimated and reference spectra of grass. The corresponding SA metrics are calculated and presented in the figure.}
    \label{fig:estimate spectra grass}
\end{figure*}

\subsection{{Comparison on Relighting}}
Further evaluation is conducted on the downstream task of relighting to assess the power of the reconstructed HSIs. Relighting is a crucial image processing task aimed at removing the influence of the illuminant in the scene and recovering the true color of objects. The color captured by the RGB camera is determined by the physical interaction between light and surface reflectance. The RGB-based relighting methods can only model the light-reflectance interaction in three channels, issuing in intrinsic errors \cite{ZHANG2024127474, 773963}. In contrast, HSI can offer broader wavelengths and more refined light-reflectance interaction modeling, making it more suitable for relighting \cite{7274900}.

In this experiment, we treat the HSIs provided by the dataset as reflectance images, and the ground truth color of the objects is rendered under CIE D65 illuminant. CIE D65 represents the average daylight with a color temperature of around 6500 K and is typically served as a standard illuminant condition \cite{Yang2010}.
The input RGB images are lighted by different illuminants, including CIE A and CIE F6, and then generated using the SSF provided by the NTIRE22 dataset. 
For the HSI-based relighting, we first estimate the lighted HSI from the input RGB image using the SR method. Next, we remove the illuminant in the lighted HSI and relight it under CIE D65 illuminant. The output RGB image is subsequently rendered using the SSF. All SR models used are trained on the NTIRE22 dataset without considering the effect of illuminant, and evaluated on the CAVE, ICVL, and NTIRE22 datasets. 
For the RGB-based relighting (RGB-re) the illuminant is projected into the RGB space via the SSF. Then the reflectance image estimation and relighting are performed directly in the RGB space. Detailed explanations of the HSI-based relighting and RGB-re can be found in the Appendix. 

In addition to RGB-re, which requires the input and ground truth illuminants, we also introduce two widely used automatic white balance (AWB) algorithms, namely the gray world (GW) and the perfect reflector (PR) \cite{AWB} as additional baselines for relighting in the RGB space. GW assumes the average color of the scene to be gray under the illuminant. PR assumes the existence of at least one perfect reflector in the scene, which reflects all incoming light. These assumptions enable AWB methods to automatically estimate the illuminant and adjust the scene's color.

\begin{figure}
    \centering
    \includegraphics[width=0.98\columnwidth]{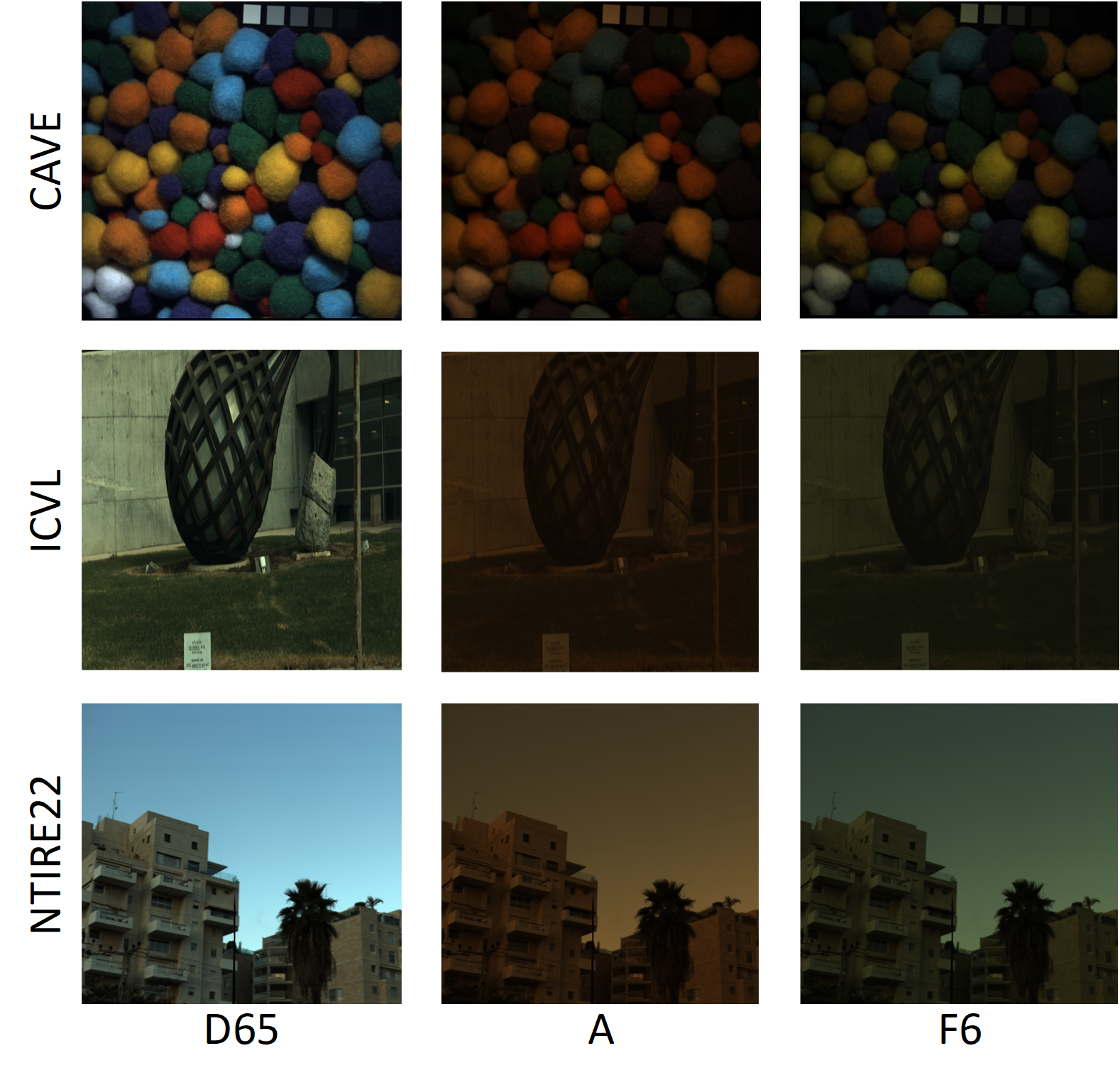} % Reduce the figure size so that it is slightly narrower than the column.
    \caption{Illustration of RGB images lighted by different illuminants. The first column is lighted by CIE D65 (considered as the ground truth), and the rest are lighted by different illuminants. From top to bottom, the scenes are from the CAVE, ICVL, and NTIRE22 datasets, respectively.
    }
    \label{fig:illuminant gt}
\end{figure}

Some ground truth and input RGB images are illustrated in Fig. \ref{fig:illuminant gt}. CIE A illuminant produces a yellowish color, and CIE F6 illuminant gives a greenish color.
The PSNRs averaged across the CAVE, ICVL, and NTIRE22 datasets are elucidated in Table \ref{tab:light}.
The performance of methods relighting in the RGB space is not satisfactory in most cases. Particularly for the AWB methods, as they heavily rely on the assumptions of the scene. With the ground truth illuminant granted, the RGB-re performance is better than the AWB methods but is restricted by the coarse light-reflectance interaction modeling. For HSI-based methods, while MST++ achieves high SR performance, as presented in Table \ref{tab:results}, its performance on relighting is less satisfactory. This may be due to its limited generalization ability across various illuminants. R2HGAN performs second-best in most cases, benefiting from the GAN-based training strategy. Our method consistently achieves the best performance in most cases.

\begin{figure*}
    \centering
    \includegraphics[width=0.98\textwidth]{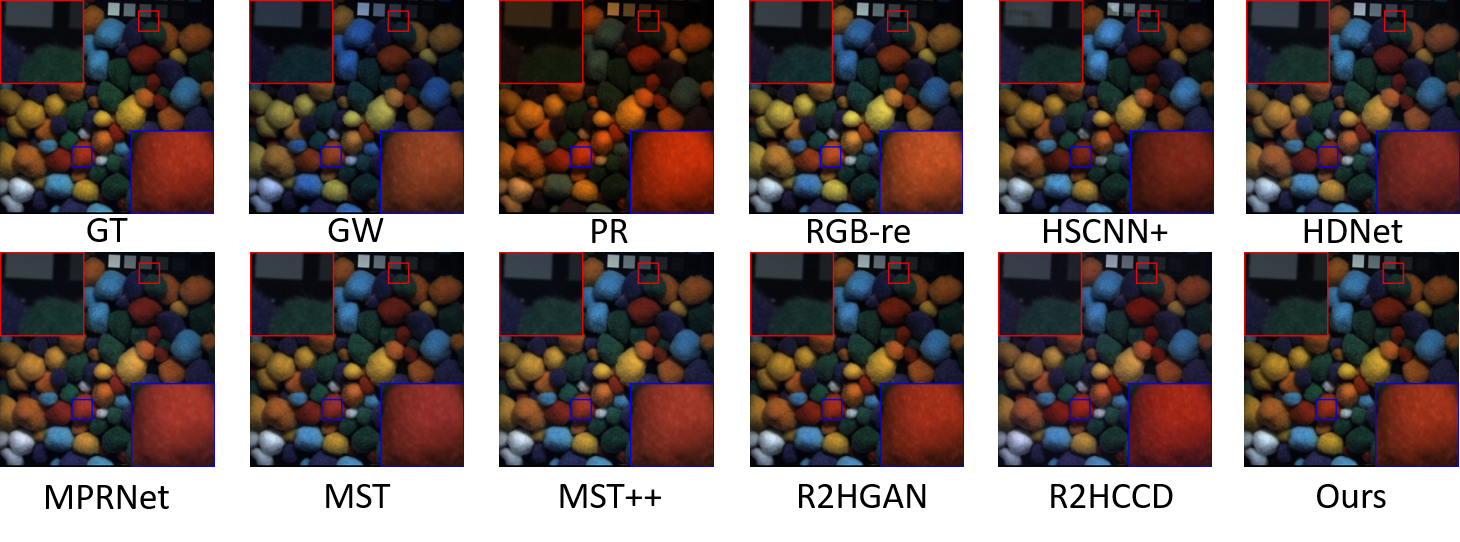} % Reduce the figure size so that it is slightly narrower than the column.
    \caption{Illustration of the ground truth and relighting results under CIE A illuminant on the CAVE dataset.
    }
    \label{fig:CAVE relgiht a1}
\end{figure*}

\begin{figure*}
    \centering
    \includegraphics[width=0.98\textwidth]{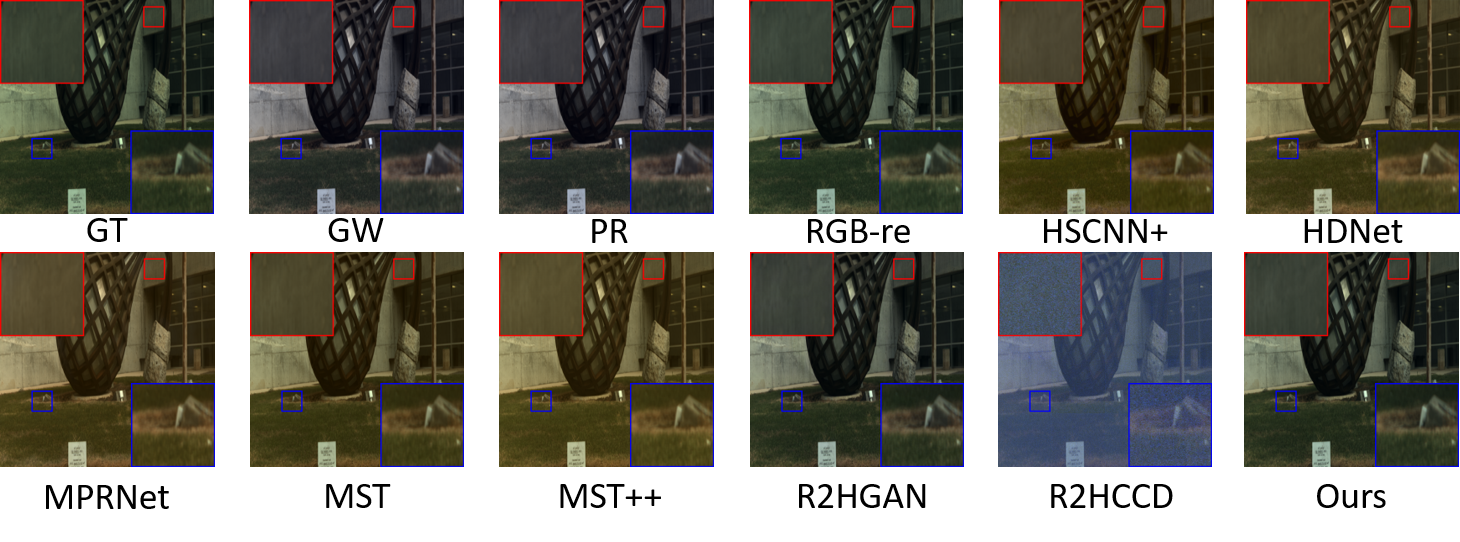} % Reduce the figure size so that it is slightly narrower than the column.
    \caption{Illustration of the ground truth and relighting results under CIE A illuminant on the ICVL dataset.
    }
    \label{fig:ICVL relight a1}
\end{figure*}

\begin{figure*}
    \centering
    \includegraphics[width=0.98\textwidth]{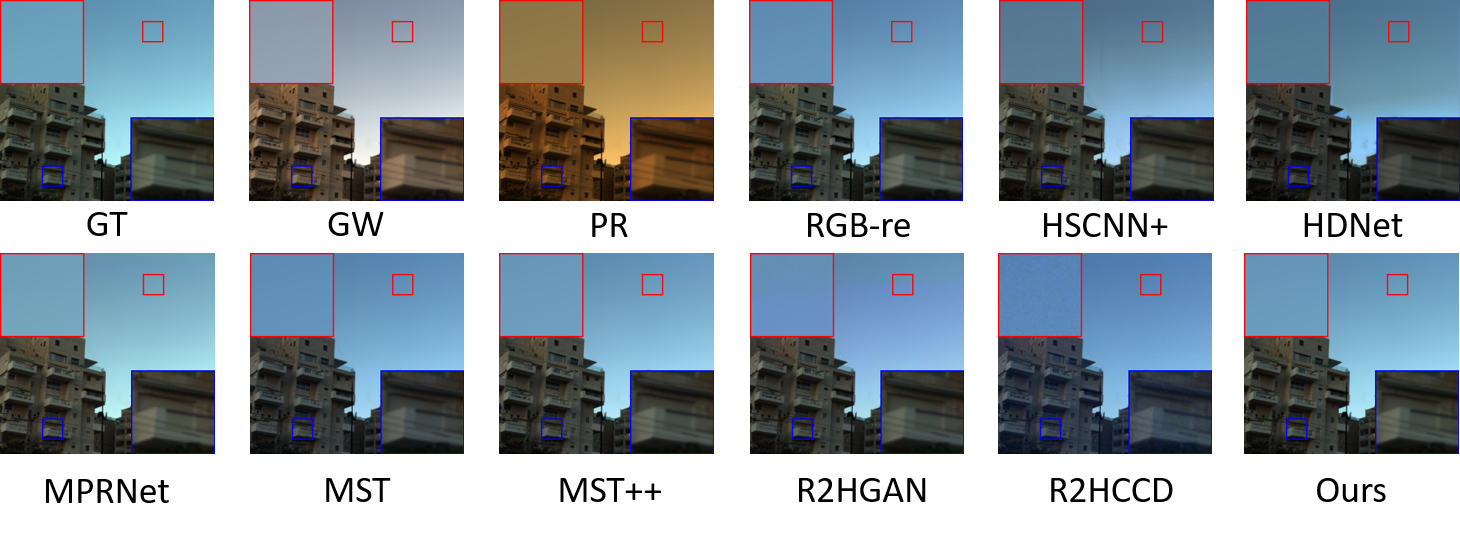} % Reduce the figure size so that it is slightly narrower than the column.
    \caption{Illustration of the ground truth and relighting results under CIE A illuminant on the NTIRE22 dataset.
    }
    \label{fig:NTIRE relight a1}
\end{figure*}

The visual results for relighting under CIE A are shown in Fig. \ref{fig:CAVE relgiht a1}--\ref{fig:NTIRE relight a1}. For the relighting results under CIE F6, please refer to the Supplementary Material.
Obvious color shifts are observed in most cases for methods relighting in the RGB space. RGB-re brings on a yellowish color in Fig. \ref{fig:CAVE relgiht a1}. The color seriously shifts to gray for GW and PR in Fig. \ref{fig:ICVL relight a1}. PR apparently fails to recover the color of the sky in Fig. \ref{fig:NTIRE relight a1}. 
For the HSI-based relighting, in Fig. \ref{fig:CAVE relgiht a1}, it is evident that HSCNN+ introduces spatial artifacts in the flattened regions. The red color rendered by alternative methods such as HDNet, MST, and MST++ appears darker than the ground truth. In Fig. \ref{fig:ICVL relight a1}, the colors of grass and wall recovered by discriminative methods exhibit noticeable errors. Additionally, the RGB images relighted by R2HCCD shift significantly to blue. In Fig. \ref{fig:NTIRE relight a1}, the color of sky produced by HSCNN+, HDNet, MST, and R2HCCD are deeper than the ground truth. R2HGAN shows satisfactory performance but introduces artifacts in the sky. In summary, the color reconstructed by our method demonstrates high fidelity and is consistent with the ground truth, validating the effectiveness of our method in relighting.


\begin{table}
    \caption{Quantitative comparison of the proposed method with alternative methods on relighting under different illuminants. The best results are \best{bold}, and the second-best results are \subbest{underscored}.}
      \centering
      \begin{tabular}{c|cc|cc|cc}
      \toprule
      \multirow{2}{*}{Method}  & \multicolumn{2}{c|}{CAVE} & \multicolumn{2}{c|}{ICVL} & \multicolumn{2}{c}{NTIRE22} \\
       & A & F6 & A & F6 & A & F6 \\
      \midrule
      GW                                        & 33.67 & 34.36 & 32.53 & 32.44 & 29.95 & 29.76 \\
      PR                                        & 24.51 & 31.41 & 22.83 & 27.57 & 21.02 & 27.40 \\
      RGB-re                                    & 38.62 & 42.89 & 38.37 & \best{42.16} & 38.24 & 38.46 \\
      \midrule
      HSCNN+                                    & 31.05 & 30.98 & 29.25 & 28.99 & 28.30 & 29.40 \\
      HDNet                                     & 38.81 & 39.42 & 32.56 & 34.19 & 32.31 & 36.21 \\
      MPRNet                                    & 38.96 & 38.59 & 30.64 & 33.14 & 35.06 & 36.82 \\
      MST                                       & 41.03 & 41.32 & 31.98 & 35.13& 37.81 & 40.52 \\
      MST++                                     & 40.57 & \subbest{43.26} & 30.45 & 32.68 & 37.21 & 40.16 \\
      \midrule
      R2HGAN                                    & \subbest{41.52} & 39.76 & \subbest{39.98} & \subbest{40.48} & \subbest{42.40} & \subbest{41.18} \\
      R2H-CCD                                   & 29.40 & 34.92 & 15.70 & 19.50 & 20.36 & 31.67\\
      Ours                                      & \best{48.28} & \best{45.59} & \best{40.83} & {40.32} & \best{45.13} & \best{42.39} \\
      \bottomrule
      \end{tabular}
      \label{tab:light}
\end{table}
\subsection{{Analysis of Important Factors}}
Some important factors in our method are analyzed in this section. All the results are averaged on the NTIRE22 dataset.
\begin{table*}
  \caption{Study of the architecture of the proposed method. The best results are \best{bold}.}
    \centering
    \begin{tabular}{c|ccc|ccc}
    \toprule
    Architecture              & Use autoencoder & Separate unobservable feature & Autoencoder's dimension & PSNR & SSIM & SAM \\
    \midrule
    % RGB only            & w/o& w/o                           & 37.17 & 0.9788 & 5.636 \\
    Pixel DM            && &              & 32.65 & 0.9223 & 16.54 \\
    LDM             &\checkmark& & 3                    & 33.47 & 0.9059 & 7.081 \\
    Ours                &\checkmark& \checkmark& 3                   & \best{42.26} & \best{0.9888} & \best{4.355} \\
    ULDM (6D)         &\checkmark& \checkmark& 6                   & 41.34 & 0.9890 & 4.699 \\
    \bottomrule
    \end{tabular}
    \label{tab:ablation}
\end{table*}

\subsubsection{Study of Architecture}
Ablation studies are conducted to investigate the effectiveness of the proposed architecture. The experimental settings and results are presented in Table \ref{tab:ablation}. The Pixel DM, which represents the standard DM implemented in the pixel space, exhibits the poorest performance, emphasizing that the autoencoder is crucial for estimating the distribution of HSIs by reducing the data dimensionality. Compared to the LDM which does not separate the unobservable feature but directly encodes the entire HSI, the proposed method demonstrates a significant improvement in reconstruction performance, confirming the effectiveness and necessity of the unobservable feature separation. Moreover, the 3D unobservable feature version of the proposed method outperforms the 6D one, suggesting that the 3D manifold is sufficient for unobservable feature modeling, and the increasing dimensionality of the manifold introduces unnecessary complexity, which hampers the estimation process.

\subsubsection{Study of Alignment Loss Weight $\lambda$}
The impact of the alignment loss weight parameter $\lambda$ is also investigated in Table \ref{tab:lambda}. The test values of $\lambda$ are set to 0, 0.1, and 0.2. The results indicate that the best performance is achieved when $\lambda$ is set to 0.1. The parameter $\lambda$ serves as a regularization term, guiding the SpeUAE to learn an unobservable feature manifold aligned with the RGB space, thus enabling the SpaAE of RGB-LDM to manipulate the spatial structures accurately. A smaller $\lambda$ weakens the regularization, leading to less accurate alignment, while a larger $\lambda$ can cause the model to converge to a local optimum in spectral structure representation.
Interestingly, even when $\lambda=0$, our method still achieves acceptable.
This is because the SpaAE does not provide spectral information, and any discrepancy between $\hat{\Fcal}^\text{Un}$ and $\Fcal^\text{Un}$ in $\mathcal{L}_\text{align}$ results in spectral information loss which may further lower the SR performance in $\mathcal{L}_\text{HSI-Re}$. As a result, $\mathcal{L}_\text{align}$ is prone to decrease even when $\lambda = 0$. Nevertheless, while learning an RGB-like manifold is possible with $\lambda = 0$, the absence of regularization makes it challenging for the SpeUAE to accurately align with the spectral manifold, resulting in inferior SR performance to $\lambda=0.1$.

\begin{table}
    \caption{Study of the weight parameter $\lambda$. The best results are \best{bold}.}
      \centering
      \begin{tabular}{c|ccc}
      \toprule
      $\lambda$ & PSNR & SSIM & SAM \\
      \midrule
      0                              & 41.84 & \best{0.9895} & 4.638 \\
      0.1                            & \best{42.26} & {0.9890} & \best{4.355} \\
      0.2                            & 41.77 & 0.9889 & 4.754 \\
      \bottomrule
      \end{tabular}
      \label{tab:lambda}
  \end{table}

\subsubsection{Study of the Number of Diffusion Steps}
\begin{figure}
    \centering
    \includegraphics[width=0.98\columnwidth]{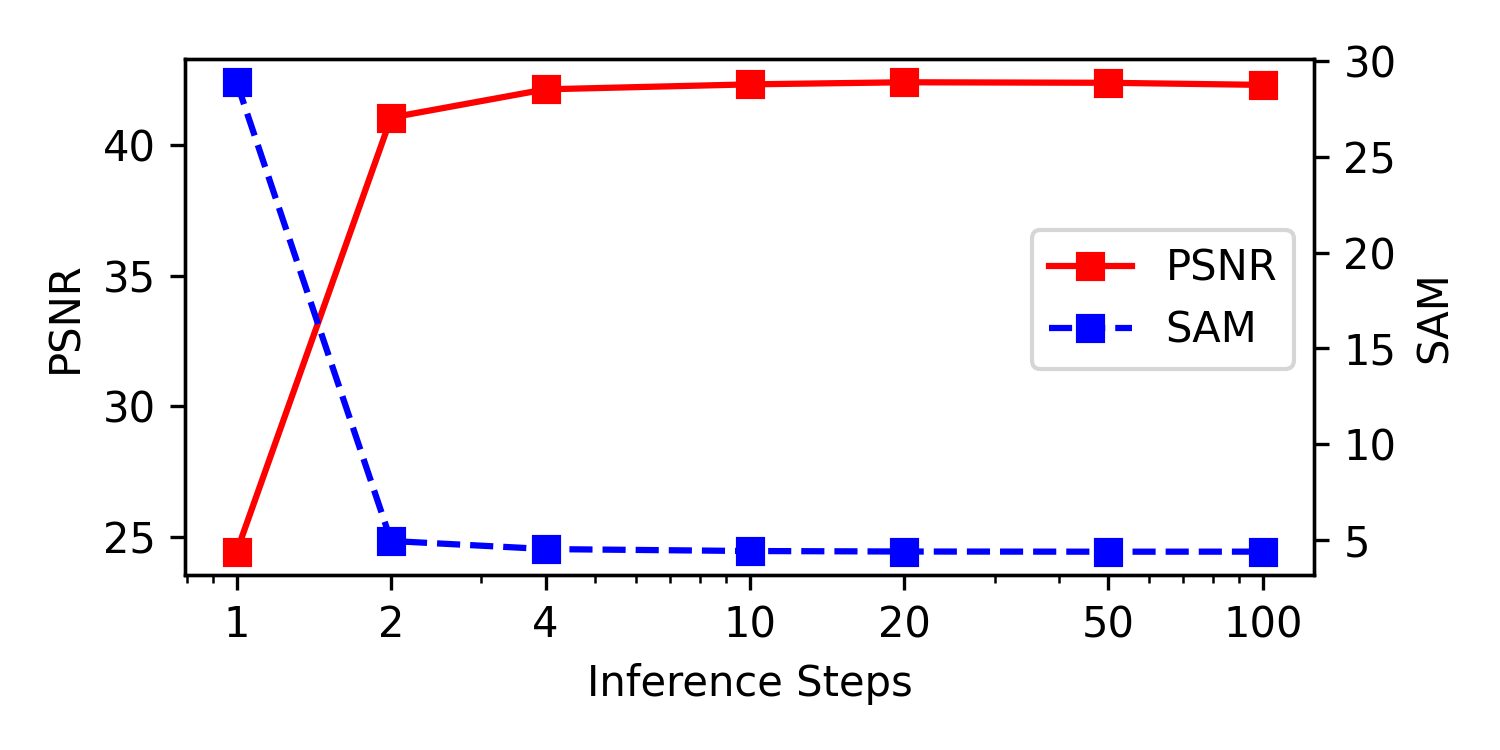} % Reduce the figure size so that it is slightly narrower than the column.
    \caption{Study of the number of diffusion steps in terms of PSNR and SSIM.}
    \label{fig:ablation}
\end{figure}

The number of diffusion steps during inference using the DDIM sampler is assessed, varying from 1 to 100. As shown in Fig. \ref{fig:ablation}, the reconstruction performance improves and then plateaus when the number of diffusion steps reaches around 10, even though both the RGB-LDM and the ULDM are trained with 1000 steps. This suggests that the SR process can be effectively accelerated by reducing the number of diffusion steps while maintaining comparable performance. To strike a balance between reconstruction accuracy and computational efficiency, we recommend using 20 diffusion steps in practice.

\section{Conclusion}
We introduce a novel generative SR method named ULDM that promotes the estimation of the unobservable feature capitalizing on the rich spatial knowledge in the RGB-LDM. The RGB-LDM is converted to the ULDM by a two-stage training pipeline which includes spectral structure representation learning and spectral-spatial joint distribution learning. In the first stage, the SpeUAE is trained to extract and compress the spectral unobservable feature into the 3D manifold. The alignment loss is employed to ensure the similarity between the 3D manifold and the RGB space, which permits the SpaAE of RGB-LDM to represent the spatial structure. The decoupling of spectral-spatial structure representation and the power of the SpaAE efficiently produces a compact and robust latent space for spectral-spatial joint distribution learning. In the second stage, the unobservable feature is embedded into this latent space, and the ULDM is derived by fine-tuning the RGB-LDM in the latent space via denoising the noisy unobservable features with guidance from the corresponding RGB images. During inference, the ULDM estimates the unobservable feature corresponding to the input RGB image. Subsequently, the SpeUAE reconstructs the HSI based on the RGB image and the estimated feature. Experimental results on SR and relighting highlight its excellent ability to reconstruct spectral information and adapt to various downstream applications.

{\appendix[RGB-based and HSI-based Relighting]
\label{app:relighting}
The color captured by the RGB camera is determined by the light-reflectance interaction and the SSF of the camera. This physical imaging process can be modeled as
\begin{align}
    \label{eq:illuminant image}
    \Ical^\text{RGB} &= \Rcal \times_1 \text{diag}(\lbf) \Pbf^T \\
    \label{eq:HSI illuminant}
    &= \Ical^\text{HS} \times_1 \Pbf^T
\end{align}
where $\lbf\in\Rbb^{B}$ is the illuminant, $\Rcal\in\Rbb^{B\times N_x\times N_y}$ is the reflectance, $\text{diag}(\cdot)$ is a diagonal matrix with the vector $\cdot$ on the diagonal.

For typically RGB-based relighting methods, as the whole spectrum is not available, the imaging process is generally down-sampled to three channels as
\begin{align}
    \label{eq:RGB illuminant}
    \Ical^\text{RGB} = \Rcal^\text{down}\times_1 \text{diag}(\lbf^\text{down}) 
    % \label{eq:HSI illuminant}
\end{align}
Here, to distinguish from the full-spectrum illuminant and reflectance, we denote the down-sampled illuminant and reflectance used in RGB-based relighting as $\lbf^\text{down}\in\Rbb^{3}$ and $\Rcal^\text{down}\in\Rbb^{3\times N_x\times N_y}$, respectively. Comparing Eq. (\ref{eq:RGB illuminant}) with Eq. (\ref{eq:illuminant image}), we have $\lbf^\text{down} = \Pbf \lbf$ and $\Rcal^\text{down} = \Rcal\times_1 \Pbf^T$. 

Therefore, given the input RGB image $\Ycal^\text{RGB}$, and input illuminant $\lbf$, the RGB-re can be formulated as
\begin{align}
    \label{eq:RGB relighting}
    \hat{\Ical}^\text{RGB} =  \Ycal^\text{RGB} \times_1 \text{diag}(\Pbf \lbf)^{-1} \text{diag}(\Pbf \lbf^\text{D65})
\end{align}
where $\lbf^\text{D65}$ is CIE D65 illuminant. The down-sampling of illuminant and reflectance restricts the capability of modeling the light-reflectance interaction and brings on intrinsic errors.

For the HSI-based relighting, we first estimate the lighted HSI $\Ycal^\text{HS}$ from $\Ycal^\text{RGB}$ using SR methods,
\begin{align}
    \hat{\Ycal}^\text{HS} = \text{SR}(\Ycal^\text{RGB})
\end{align}
where $\text{SR}(\cdot)$ refers to one of the SR methods. Then, we remove the illuminant in $\hat{\Ycal}^\text{HS}$ and obtain the reflectance image $\hat{\Rcal}$,
\begin{align}
    \label{eq:reflectance}
    \hat{\Rcal} = \hat{\Ycal}^\text{HS} \times_1 \text{diag}({\lbf})^{-1}
\end{align}
Here, we assume that $\lbf$ is known, as there are many methods to estimate the illuminant from HSIs \cite{Glatt_2024_WACV, Zheng_2015_CVPR}. Subsequently, we render the HSI by $\hat{\Rcal}$ and $\lbf^\text{D65}$,
\begin{align}
    \hat{\Ical}^\text{HS} = \hat{\Rcal} \times_1 \text{diag}(\lbf^\text{D65})
\end{align}
Finally, we generate the output RGB image $\hat{X}^\text{RGB}$ by the SSF,
\begin{align}
    \hat{\Ical}^\text{RGB} = \hat{\Ical}^\text{HS} \times_1 \Pbf^T
\end{align}
However, for some poor illumination conditions, the power of the illuminant is nearly zero, which will lead to information loss and numerical instability to Eq. (\ref{eq:reflectance}). To avoid this, we set a threshold $\epsilon$ to the illuminant, and use linear interpolation to estimate the loss reflectance. We obtain the missing wavelength index set $\Omega = \{b|l_{b} < \epsilon\}$, and interpolate the reflectance in these wavelengths by the nearest non-zero wavelengths. The pseudocode of the relighting process is shown in Algorithm \ref{alg:relighting}. The threshold $\epsilon$ is recommended to be $10\%$ of the maximum value of the illuminant $\lbf$.
\begin{algorithm}
\caption{HSI-based Relighting}
\label{alg:relighting}
\begin{algorithmic}[1]
\REQUIRE RGB image $\Ycal^\text{RGB}$, SSF $\Pbf$, illuminant $\lbf$, CIE D65 illuminant $\lbf^\text{D65}$, SR method $\text{SR}(\cdot)$, threshold $\epsilon$
\ENSURE Relighted RGB image $\hat{X}^\text{RGB}$
\STATE Estimate the lighted HSI $\hat{\Ycal}^\text{HS} = \text{SR}(\Ycal^\text{RGB})$
\STATE Remove the illuminant in $\hat{\Ycal}^\text{HS}$ and obtain the reflectance image $\hat{\Rcal}$ by Eq. (\ref{eq:reflectance})
\STATE Obtain the index set $\Omega = \{b|l_{b} < \epsilon\}$
\FOR{$n = 1$ to $N_x$}
\FOR{$m = 1$ to $N_y$}
\STATE $\hat \Rcal_{:, n,m} = \text{interp}(\hat \Rcal_{:, n,m}, \Omega)$
\ENDFOR
\ENDFOR
\STATE Render the reflectance image $\hat{R}$ by $\lbf^\text{D65}$
\STATE Generate the output RGB image $\hat{X}^\text{RGB}$ by $\Pbf$
\RETURN $\hat{X}^\text{RGB}$
\end{algorithmic}
\end{algorithm}
}


 
\bibliographystyle{IEEEtran}

\bibliography{ref}


% \end{IEEEbiography}



\vfill

\end{document}